\documentclass[a4paper,twoside]{article}

\usepackage{epsfig}
\usepackage{subfigure}
\usepackage{calc}
\usepackage{amssymb}
\usepackage{amstext}
\usepackage{amsmath}
\usepackage{amsthm}
\usepackage{multicol}
\usepackage{pslatex}
\usepackage{apalike}
\usepackage{url}
\usepackage{bm}
\usepackage[normalem]{ulem}
\usepackage{epstopdf}
\usepackage{SCITEPRESS}     

\begin{document}

\title{Using Apache Lucene to Search Vector of Locally Aggregated Descriptors}


\author{\authorname{Giuseppe Amato, Paolo Bolettieri, Fabrizio Falchi, Claudio Gennaro, Lucia Vadicamo }
    \affiliation{ISTI-CNR, Via G. Moruzzi, 1, 56124, Pisa, Italy}
   \email{\{giuseppe.amato, paolo.bolettieri, fabrizio.falchi, claudio.gennaro, lucia.vadicamo\}@isti.cnr.it}
}


\abstract{
Surrogate Text Representation (STR) is a profitable solution to efficient similarity search on metric space using
conventional text search engines, such as Apache Lucene. This technique is
based on comparing the permutations of some reference objects in place of the original metric distance.
However, the Achilles heel of STR approach is the need to reorder the result set of the search according to the metric distance.
This forces to use a support database to store the original objects, which requires efficient random I/O on a fast secondary memory (such as flash-based storages). 
In this paper, we propose to extend the Surrogate Text Representation to specifically address a class of visual metric objects known as Vector of Locally Aggregated Descriptors (VLAD).
This approach is based on representing the individual sub-vectors forming the VLAD vector with the STR, providing a finer
representation of the vector and enabling us to get rid of the reordering phase.
The experiments on a publicly available dataset show that the extended STR outperforms the baseline STR achieving satisfactory performance near to the one obtained with the original VLAD vectors.}

\onecolumn \maketitle \normalsize \vfill

\section{Introduction}
\label{sec:introduction} 

\noindent Multimedia information retrieval on a large scale database has to address 
at the same time both issues related to effectiveness and efficiency.
Search results should be pertinent to the submitted queries, and
should be obtained quickly, even in presence of very large multimedia
archives and simultaneous query load.

Vectors of Locally Aggregated Descriptors (VLAD) \cite{jegou12}
were recently proposed as a way of producing compact
representation of local visual descriptors, as for instance SIFT
\cite{lowe04}, while still retaining high level of accuracy. In
fact, experiments, demonstrated that VLAD accuracy is higher than
Bag of Words (BoW) \cite{sivic03}. The advantage of BoW representation is that
it is very sparse and allows using inverted files to also achieve
high efficiency. VLAD representation is not sparse, so general
indexing methods for similarity searching \cite{metric-book} must
be used, which are typically less efficient than inverted files.

One of the best performing generic methods for similarity
searching, is the use of permutation based indexes
\cite{Navarro2008,mtap12}. Permutation based indexes rely on
the assumption that to objects that are very similar, ``see'' the
space around them in a similar way. This assumption is exploited
by representing the objects as the ordering of a fixed set of
reference objects (or pivots), according to their distance from
the objects themselves. If two objects are very similar, the two
corresponding ordering of the reference objects will be similar as well.

However, measuring the similarity between objects using the
similarity between permutations is a coarse approximation. In
fact, in order to achieve also high accuracy, similarity between
permutations is used just to identify an appropriate set of
candidates, which is then reordered according to the original
similarity function to obtain the final result. This reordering
phase, contributes to the overall search cost.

Given that objects are represented as ordering (permutations) of
reference objects, permutation based indexes offer the possibility of using
inverted files, in every similarity searching problem, where
distance functions are metric functions. In fact, \cite{ecdl10}
presents an approach where the Lucene text search engines, was
used to index and retrieve objects by similarity. The technique is
based on an encoding of the permutations by means of a
\emph{Surrogate Text Representation} (STR). In this respect, VLAD
can be easily indexed using this technique, as discussed in
\cite{Amato:2014:IVL:2578726.2578788} so that efficient and
effective image search engines can be built on top of a standard text
search engine.

In this paper, we propose an advancement on this basic techniques,
which exploits the internal structure of VLAD. Specifically, the
STR technique is applied, independently, to portions of the entire
VLAD. This leads, at the same time, to higher efficiency and
accuracy without the need of executing the reordering of the set of
candidates, which was mentioned above. The final result is
obtained by directly using the similarity between the permutations
(the textual representation), so saving both time in the searching
algorithms, and space, since the original VLAD vectors no longer
need to be stored.

The paper is organized as follows. Section \ref{sec:related-work}
makes a survey of the related works. Section \ref{sec:vlad} provides a brief introduction to the VLAD approach. Section \ref{sec:proposed
approach} introduces the proposed approach. Section
\ref{sec:experiments} discusses the validation tests. Section
\ref{sec:conclusions} concludes.

\section{Related Work}
\label{sec:related-work} \noindent

\noindent
In the last two decades, the breakthroughs in the field of image retrieval have been mainly based on the use of the local features.
Local features, as SIFT \cite{lowe04} and SURF
\cite{bay06}, are visual descriptors of selected interest points of an image. Their use allows one to effectively match local structures between images.
However, the costs of comparison of the local features lay some limits on large scale, since each image is represented by typically thousands
of local descriptors.
%
Therefore, various methods for the
aggregation of local features have been proposed. 


One of the most popular aggregation method is the \emph{Bag-of-Word} (BoW), initially proposed in \cite{sivic03,csurka04} for matching object in videos. BoW uses a \textit{visual vocabulary} to quantize the local descriptors extracted from images; each image is then represented by a histogram of occurrences of visual words.
The BoW approach used in computer vision is very similar to the BoW used in natural language processing and information retrieval \cite{salton86}, thus many text indexing techniques, as inverted files \cite{witten99}, have been applied for image search.
From the very beginning \cite{sivic03} words reductions techniques have been used and images have been ranked using the standard \emph{term frequency-inverse document frequency} ({tf-idf}) \cite{salton86} weighting.
In order to improve the efficiency of BoW, several approaches for the reduction of visual words have been investigated \cite{thomee10,amato13:onReducing}.
Search results obtained using BoW in CBIR (Content Based Image Retrieval) has also been improved by exploiting additional geometrical information \cite{philbin07,perdoch09,tolias11:SpeededUp,zhao13} and applying re-ranking approaches \cite{philbin07,jegou08,chum07,tolias13:queryExp}.
The baseline BoW encoding is affected by the loss of information about the original descriptors due to the quantization process. For example, corresponding descriptors in two images may be assigned to different visual words. To overcome the quantization loss, more accurate representation of the original descriptors and alternative encoding techniques have been used, such as \emph{Hamming Embedding} \cite{jegou08,Jegou2010}, \emph{soft-assignment} \cite{philbin08,vanGemert08,vanGemert10}, \emph{multiple assignment} \cite{Jegou2010,jegou10:AccurateImage},  \emph{locality-constrained linear coding} \cite{wang10}, \emph{sparse coding} \cite{yang09,boureau10} and the use of \emph{spatial pyramids} \cite{lazebnik06}.

Recently, other aggregation schemes, such as the \emph{Fisher Vector} (FV) \cite{perronnin07,jaakkola98} and the \emph{Vector of Locally Aggregated Descriptors} (VLAD) \cite{JegouCVPR2010}, have attracted much attention because of their effectiveness in both image classification and large-scale image search. Both FV and VLAD use some statistics about the distribution of the local descriptors in order to transform an incoming set of descriptors into a fixed-size vector representation.

The basic idea of FV is to characterize how a sample of descriptors deviates from an average distribution that is modeled by a parametric generative model. The Gaussian Mixture Model (GMM) \cite{mclachlan2000}, estimated on a training set, is typically used as generative model and might be understood as a ``probabilistic visual vocabulary''.

While BoW counts the occurrences of visual words and so takes in account just 0-order statistics, the VLAD approach, similarly to BoW, uses a visual vocabulary to quantize the local descriptors of an image. The visual vocabulary is learned using a clustering algorithm, as for example the $k$-means. Compared to BOW, VLAD exploits more aspects of the distribution of the descriptors assigned to a visual word. In fact, VLAD encodes the accumulated difference between the visual words and the associated descriptors, rather than just the number of descriptors assigned to each visual word.
As common post-processing step VLAD is power and L2 normalized \cite{jegou12,Perronnin2010}. Furthermore, PCA
dimensionality reduction and product quantization have been
applied and several enhancements to the basic VLAD have been
proposed \cite{arandjelovic13:allAbVALD,Perronnin2010,chen11,delhumeau13,zhao13}

In this work, we will focus on VLAD which is very similar to FV. In fact VLAD has been proved to be a simplified non-probabilistic version of FV that performs very similar to FV \cite{jegou12}.
However, while BoW is a sparse vector of occurrence, VLAD is not. Thus, inverted files cannot be directly applied for indexing and Euclidean Locality-Sensitive Hashing \cite{Datar:2004} is, as far as we know, the only technique tested with VLAD.
Many other similarity search indexing techniques \cite{metric-book} could be applied to VLAD.
A very promising direction is Permutation-Based Indexing \cite{Navarro2008,mtap12,MiPai}. In particular the MI-File
allows one to use inverted files to perform similarity search with an
arbitrary similarity function.
Moreover, in \cite{ecdl10,AmatoCBMI2011}
a Surrogate Text Representation (STR) derived from the MI-File has been proposed.
The conversion of the image description in textual form enables us to exploit
the off-the-shelf search engine features with a little implementation effort.

In this paper, we extend the STR approach to deal with the VLAD descriptions
comparing both effectiveness and efficiency with the STR baseline approach, which has been studied in \cite{amato13}.
The experimentation was carried out on the same hardware and software infrastructure using
a publicly available INRIA Holidays \cite{jegou08} dataset and comparing the effectiveness with the sequential scan.

\section{Vector of Locally Aggregated Descriptors (VLAD)}\label{sec:vlad}
\noindent
The VLAD representation was proposed in \cite{Jegou2010}. As for
the BoW, a visual vocabulary, here called \emph{codebook},
$\{\bm{\mu}_1, \ldots,
\bm{\mu}_{\mathcal{K}}\}$\footnote{Throughout the paper bold
letters denote row vectors.} is first learned using a cluster
algorithm (e.g. $k$-means). Each local descriptor $\mathbf{x}_t$
is then associated with its nearest visual word (or
\emph{codeword}) $NN(\mathbf{x}_t)$ in the codebook. For each
codeword the differences between the sub-vectors $\mathbf{x}_t$
assigned to $\bm\mu_i$ are accumulated:
$$
\mathbf{v}_i=\sum_{\mathbf x_t:NN(\mathbf x_t)=i} \mathbf x_t - \bm{\mu}_i
$$
The VLAD is the concatenation of the accumulated sub-vectors, i.e.
$\mathbf V=(\mathbf v_1, \ldots, \mathbf v_\mathcal{K})$.  Throughout the paper, we refer to the accumulated sub-vectors $\mathbf{v}_i$ simply as ``sub-vectors''.

Two normalization are performed: first, a power normalization with power $0.5$; second, a 
L2 normalization. After this process two descriptions can be compared using the inner product.

The observation that descriptors are relatively sparse and very
structured suggests performing a principal component analysis
(PCA) to reduce the dimensionality of the VLAD. In this work, we
decide not to use dimensionality reduction techniques because we
will show that our space transformation approach is independent
from the original dimensionality of the description. In fact, the
STR approach that we propose, transforms the VLAD description in a
set of words from a vocabulary that is independent from the
original VLAD dimensionality. 

\section{Surrogate Text Representation for VLAD Vectors}
\label{sec:proposed approach} \noindent
In this paper, we propose to index VLAD using
a text encoding that allows using any text retrieval engine to
perform image similarity search.
As discussed later, we implemented this idea on top of
the Lucene text retrieval engine\footnote{http://lucene.apache.org}.

To this end, we extend the permutation-based approach developed by Chavez et al. \cite{Navarro2008} to deal with the internal representation of the VLAD vectors. In this section, we first introduce the basic principle of the permutation-based approach and then describe the generalization to VLAD vectors.

\subsection{Baseline Permutation-based Approach and Surrogate Text Descriptor}

The key idea of the Permutation-based approach relies on the
observation that if two objects are near one another, they have a
similar view of the objects around them. This means that the
orderings (permutations) of the surrounding objects, according to
the distances from the two objects, should be similar as well.

Let $\mathcal{D}$ be a domain of objects (features, points, etc.),
and $d:\mathcal{D}\times\mathcal{D}\rightarrow \mathbb{R}$ a
distance function able to assess the dissimilarity between two
objects of $\mathcal{D}$. Let $R\subset\mathcal{D}$, be a set of
$m$ \emph{distinct} objects (reference objects), i.e.,
$R=\{r_1,\ldots,r_m\}$. Given any object $o \in \mathcal{D}$, we
denote the vector of rank positions of the reference objects,
ordered by increasing distance from $o$, as $\mathbf{p}(o) =
(p_1(o),\ldots,p_m(o))$. For instance, if $p_3(o) = 2$ then $r_3$
is the 2nd nearest object to $o$ among those in $R$. The essence
of the permutation-based approach is to allow executing similarity
searching exploiting distances between permutations in place of
original objects' distance. This, as discussed in the following, has the
advantage of allowing using a standard text retrieval engine to
execute similarity searching.

There are several standard methods for comparing two ordered lists, such as Kendall's tau distance, Spearman Footrule distance, and Spearman Rho distance. In this paper, we concentrate our attention on the latter distance, which is also used in \cite{Navarro2008}. The reason of this choice (explained later on) is tied to the way standard search engines process the similarity between documents and query.

In particular, we exploit a generalization of the Spearman Rho distance that allows us to compare two top-$k$ ranked lists. Top-$k$ list is a particular case of a partial ranked list, which is a list that contains rankings for only a subset of items. For top-$k$ lists, we can use a generalization of the Spearman Rho distance $\tilde{d}(o,q)$, called \textit{location parameter distance} \cite{FAGIN2003b}, which assigns a rank $k+1$ for all items of the list that have rank greater than $k$.

In particular, let $k$ be an integer less or equal than $m$, and $\mathbf{p}^k(o)=(p^k_1(o),\ldots,p^k_m(o))$ the vector defined as follows:
\begin{equation}
    {{p}_i^{k}}(o)=\left\{ \begin{matrix}
        p_i(o)\text{ if }p_i(o)\le k\text{ }  \\
        k+1\text{ if }p_i(o)>k  \\
    \end{matrix}. \right.\label{eq:pk}
\end{equation}

Given two top-$k$ ranked lists with $k=k_q$ and $k=k_x$, we define the approximate distance function $\tilde{d}(o,q)$ as follows:
\begin{equation} \label{eq:d} \tilde{d}(o,q)=||\mathbf{p}^{k_x}(o)-\mathbf{p}^{k_q}(q)||_2, \end{equation}
where $k_q$ is used for queries and $k_x$ for indexing. The reason for using two different $k$ relies on the fact the performance of the inverted files is optimal when the size of the queries are much smaller than the size of documents. Therefore, we will typically require that $k_q \leq k_x$. 

Since, the square root in Eq. (\ref{eq:d}) is monotonous, it does not affect the ordering \cite{FAGIN2003b}, so we can safely use $\tilde{d}(o,q)^2$ instead of its square-root counterpart:
\begin{equation} \begin{array}{ll}\label{eq:d2} \tilde{d}(o,q)^2 = \sum^m_{i=1}{{\left(p^{k_x}_i(o)-p^{k_q}_i(q)\right)}^{2}} =\\
||\mathbf{p}^{k_x}(o)||_2^2 + ||\mathbf{p}^{k_q}(q)||_2^2 - 2 \,\mathbf{p}^{k_x}(o)\cdot \mathbf{p}^{k_q}(q) \end{array}\end{equation}

\begin{figure*}[t] \centering
    \includegraphics[ width=14cm]{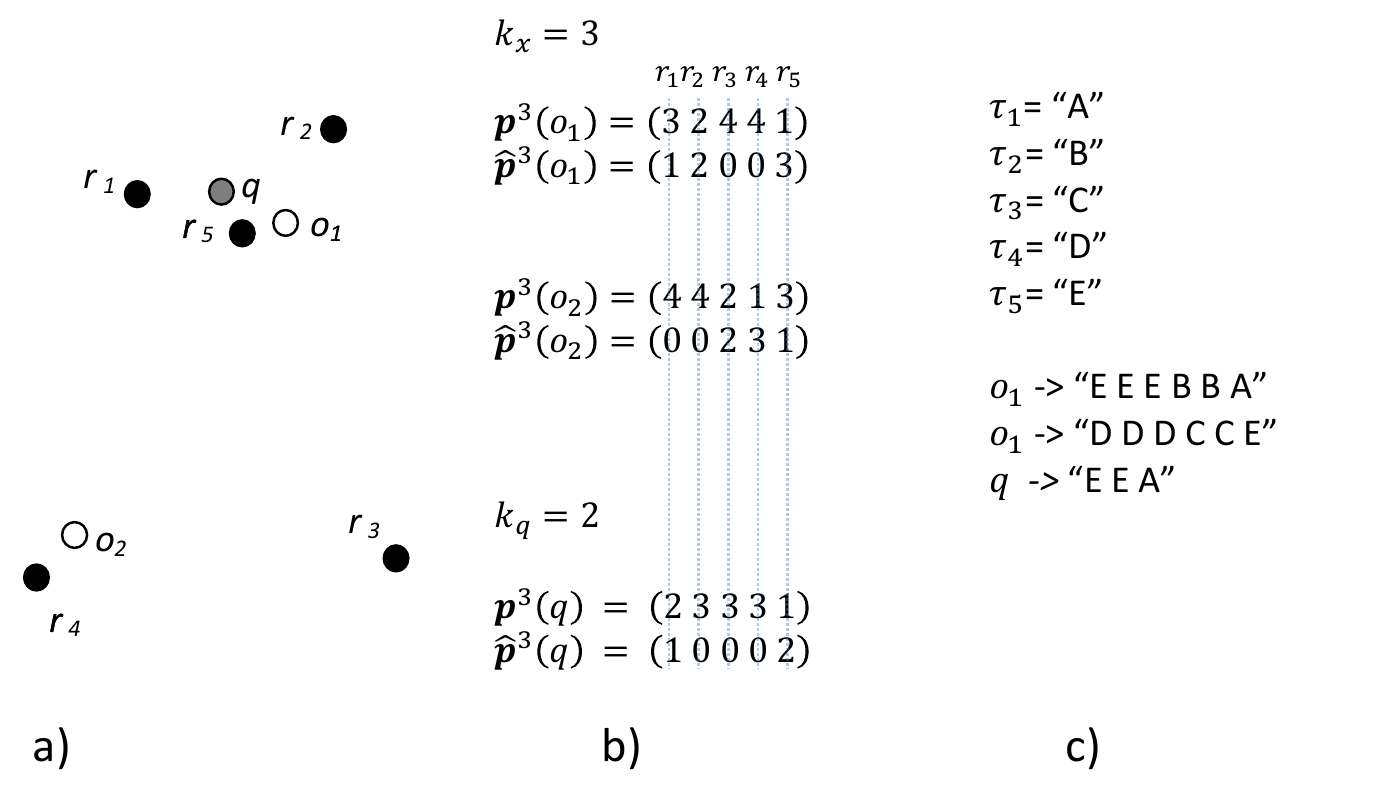}
    \caption{Example of perspective based space transformation. a)
        Black points are reference objects; white points are data objects;
        the grey point is a query. b) Encoding of the data objects in the
        transformed space. c) Encoding of the data objects in textual form.
            }
    \label{Fig:fig-trans-examp}
\end{figure*}

Figure \ref{Fig:fig-trans-examp} exemplifies the transformation
process. Figure \ref{Fig:fig-trans-examp}a sketches a number of
reference objects (black points), objects (white points), and
a query object (gray point). Figure \ref{Fig:fig-trans-examp}b
shows the encoding of the data objects in the transformed space.
We will use this illustration as a running example throughout the remainder of
the paper.

So far, we have presented a method for approximating the function $d$. However, our primary objective is to implement the function $\tilde{d}(o,q)$ in an efficient way by exploiting the built-in cosine similarity measure of standard text-based search engines based on vector space model. For this purpose, we associate each element $r_i\in R$ with a unique key $\tau_i$. The set of keys $\{\tau_1,\ldots,\tau_m\}$ represents our so-called ``reference-dictionary''.
Then, we define a function $t^k(o)$ that returns a space-separated concatenation of zero or more repetitions of ${\tau }_i$ keywords, as follows:
\({{t}^{k}}(o)=\bigcup\limits_{i=1}^{m}{\bigcup\limits_{j=1}^{k+1-p_{i}^{k}(o)\;}{{{\tau }_{i}}}}\)
where, by abuse of notation, we denote the space-separated
concatenation of keywords with the union operator $\cup$. The
function $t^k(o)$ is used to generate the Surrogate Text Representation for both indexing
and querying purposes. $k$ assumes in general the values $k_x$ for
indexing and $k_q$ for querying. For instance, consider the case
exemplified in Figure \ref{Fig:fig-trans-examp}c, and let us assume
$\tau_1=$A, $\tau_2=$B, etc. The function ${t}^{k}$ with
$k_x=3$ and $k_q=2$, will generate the following outputs
\begin{align*}
{t}^{k_x}(o_1) = & \text{  ``E E E B B A''}\\
{t}^{k_x}(o_2) = & \text{  ``D D D C C E''}\\
{t}^{k_q}(q) \,\,\,\,= & \text{  ``E E A''}
\end{align*}

As can be seen intuitively, strings corresponding to $o_1$ and $q$ are more similar to those corresponding to $o_2$ e $q$, this reflects the behavior of the distance $\tilde{d}$. However, this should not mislead the reader: our proposal is not a heuristic, the distance between the strings corresponds exactly to the distance $\tilde{d}$ between the objects, as we will prove below.

As explained above, the objective now is to force a standard text-based search engine to generate the approximate distance function $\tilde{d}$. How this objective is obtained becomes obvious by the following considerations. A text based search engine will generate a vector representation of STRs generated with $t^{k_x}(o)$ and $t^{k_q}(q)$ containing the number of occurrences of words in texts. This is the case of the simple term-frequency weighting scheme. This means that, if for instance keyword ${\tau }_i$ corresponding to the reference object $r_i$ ($1\le i\le m$) appears $n$ times, the $i$-th element of the vector will assume the value $n$, and whenever ${\tau }_i$ does not appear it will be 0.
Let $\mathbf{k_x}$ and $\mathbf{k_q}$ be respectively the constant $m$-dimensional vectors, $(k_x+1,\ldots,k_x+1)$ and $(k_q+1,\ldots,k_q+1)$, then
\begin{equation}
\begin{array}{ll}
\mathbf{\widehat{p}}^{k_x}(o)=\mathbf{k_x}-\mathbf{p}^{k_x}(o)\\
\mathbf{\widehat{p}}^{k_q}(q)=\mathbf{k_q}-\mathbf{p}^{k_q}(q)
\end{array} \label{eq:minusvects}
\end{equation}
It is easy to see that the vectors corresponding to $t^{k_x}(o)$ and $t^{k_q}(q)$, are the same of $\mathbf{\widehat{p}}^{k_x}(o)$ and $\mathbf{\widehat{p}}^{k_q}(q)$, respectively.

The cosine similarity is typically adopted to determine the similarity of the query vector and a vector in the database of the search engine, and it is defined as:
\begin{equation}
    si{{m}_{cos}}\left( o,q \right)=\frac{{{{\mathbf{\widehat{p}}}}^{{{k}_{x}}}(o)}\cdot{{{\mathbf{\widehat{p}}}}^{{{k}_{q}}}(q)}\text{ }\!\!~\!\!\text{ }}{\left\| {{{\mathbf{\widehat{p}}}}^{{{k}_{x}}}(o)} \right\|\text{ }\!\!~\!\!\text{ }\left\| {{{\mathbf{\widehat{p}}}}^{{{k}_{q}}}(q)} \right\|} \propto  {\mathbf{\widehat{p}}}^{{k}_{x}}(o)\cdot{\mathbf{\widehat{p}}}^{{k}_{q}}(q).  \label{eq:simcos}
\end{equation}
It is worth noting that ${\mathbf{\widehat{p}}}^{{k}}$ is a permutation of the $m$-dimensional vector $(1,2,\dots, k, 0,\dots, 0)$, thus its norm equals  $\sqrt{k(k+1)(2k+1)/6}$. Since $k_x$ and $k_q$ are constants, the norms of vectors ${\mathbf{\widehat{p}}}^{{k}_{x}}$ and ${\mathbf{\widehat{p}}}^{{k}_{q}}$ are constants too, therefore can be neglected during the cosine evaluation (they do not affect the final ranking of the search result).

What we are now to show is that $si{{m}_{cos}}$ can be used as a function for evaluating a similarity of two objects in place of the distance $\tilde{d}$ and it possible to prove that the first one is a order reversing monotonic transformation of the second one (they are equivalent for practical aspects). This means that if we use $\tilde{d}(o,q)$ and we take the first $k$ nearest objects from a dataset $X \subset \mathcal{D}$ (i.e, from the shortest distance to the highest) we obtain exactly the same objects in the same order if we use ${sim}_{cos}\left(o,q\right)$ and take the first $k$ similar objects (i.e., from the greater values to the smaller ones).

By substituting Eq. \eqref{eq:minusvects} into Eq. \eqref{eq:simcos}, we obtain:
\begin{equation}
\begin{array}{ll}
\displaystyle{sim}_{cos}(o,q) \propto (\mathbf{k_x}-\mathbf{p}^{k_x}(o)) \cdot (\mathbf{k_q}-\mathbf{p}^{k_q}(q)) = \\
\ \\
= \mathbf{k_x} \cdot \mathbf{k_q} - \mathbf{k_x}\cdot\mathbf{p}^{k_q}(q) - \mathbf{k_q}\cdot\mathbf{p}^{k_x}(o) + \mathbf{p}^{k_x}(o)\cdot\mathbf{p}^{k_q}(q)\\
\end{array}\label{eq:simcos2}
\end{equation}
since $\mathbf{p}^{k_x}(o)$  ($\mathbf{p}^{k_q}(q)$) include all integers numbers from $1$ to $k_x$ ($k_q$) and the remaining assumes $k_x+1$ ($k_q+1$) values, the scalar product $\mathbf{k_x}\cdot\mathbf{p}^{k_q}(q)$ ($\mathbf{k_q}\cdot\mathbf{p}^{k_x}(o)$) is constant. We can substitute the first three member in Eq. (\ref{eq:simcos2}) with a constant $L(m,k_x,k_q)$, which depends only on $m$, $k_x$, and $k_q$ as follows:
\begin{equation}
        {sim}_{cos}(o,q) \propto L(m,k_x,k_q) + \mathbf{p}^{k_x}(o)\cdot\mathbf{p}^{k_q}(q) \label{eq:simcos3}.
\end{equation}
Finally, combining Eq. (\ref{eq:simcos3}) with Eq. (\ref{eq:d2}), we obtain:
\begin{equation}
\begin{array}{ll}
{sim}_{cos}(o,q) \propto L(m,k_x,k_q) + \frac{1}{2}||\mathbf{p}^{k_x}(o)||_2^2 +\\
\\
\qquad + \frac{1}{2}||\mathbf{p}^{k_q}(q)||_2^2 - \frac{1}{2}\tilde{d}(o,q)^2. \\
\end{array}\label{eq:simcos4}
\end{equation}
%
%
Since $||\mathbf{p}^{k_x}(o)||$ and
$||\mathbf{p}^{k_q}(q)||$ depend only on the constants $m$, $k_x$, and
$k_q$, the Eq. (\ref{eq:simcos4}) proves that
${sim}_{cos}\left(o,q\right)$ is a monotonic transformation of
${\tilde{d}(o,q)}^2$ in the form $sim_{cos}=\alpha - \beta
\tilde{d}^2$. 

To summarize, given a distance function $d$, we were
able to determine an approximate distance function $\tilde{d}$, which we transformed in a similarity measure.
We proved that this similarity measure can be obtained using the STR
and that it is equivalent from the point of view of the result ranking to $\tilde{d}$. 

Note, however, that searching using directly the distance from
permutations suffers of low precision. To improve effectiveness,
\cite{mtap12} proposes to reorder the results set according to
original distance function $d$. Suppose we are searching for the
$k$ most similar (nearest neighbors) descriptors to the query. The
quality of the approximation is improved by reordering, using the
original distance function $d$, the first $c$ ($c \geq k$)
descriptors from the approximate result set at the cost of $c$
additional distance computations.

\subsection{Blockwise Permutation-based Approach}
The idea described so far uses a textual/permutation
representation of the object as whole, however, in our particular
scenario, we can exploit the fact that VLAD vector is the result
of concatenation of sub-vectors. In short, we apply
and compare the textual/permutation representation for each
sub-vector $\mathbf{v}_i$ of the whole VLAD, independently. 
We refer to this approach as Blockwise Permutation-based approach.

As we will see, this approach has the advantage of providing a finer
representation of objects, in terms of permutations, so that no
reordering is needed to guarantee the quality of the search result.

In order to decrease the complexity of the approach and since
sub-vectors $\mathbf{v}_i$ are homogeneous, we use the same set of
reference objects $R=\{r_1,\ldots,r_m\}$ to represent them as
permutations taken at random from the dataset of VLAD vectors. Let
$\mathbf{v}_i$ be the $i$-st sub-vector of a VLAD sub-vector
$\mathbf{V}$, we denote by $\mathbf{p}^{k_x}(\mathbf{v}_i)$ the
corresponding permutation vector. Given two VLAD vectors
$\mathbf{V}=(\mathbf{v}_1, \ldots, \mathbf{v}_\mathcal{K})$ and
$\mathbf{W} = (\mathbf{w}_1, \ldots, \mathbf{w}_\mathcal{K})$,
and their corresponding concatenated permutation vectors
$\mathbf{O} = (\mathbf{p}^{k_x}(\mathbf{v}_1), \ldots,
\mathbf{p}^{k_x}(\mathbf{v}_\mathcal{K}))$ and
$\mathbf{Q}= (\mathbf{p}^{k_q}(w_1), \ldots,
\mathbf{p}^{k_q}(w_\mathcal{K}))$, we generalize the
Spearman Rho distance for two vectors $\mathbf{V}$ and
$\mathbf{W}$ as follows:

\begin{equation} \label{eq:d2vlad}
\begin{array}{ll}
\tilde{d}(\mathbf{V},\mathbf{W})^2 =\sum^\mathcal{K}_{i=1}\tilde{d}(\mathbf{v}_i,\mathbf{w}_i)^2 =\\
\sum^\mathcal{K}_{i=1} ||\mathbf{p}^{k_x}(\mathbf{v}_i)-\mathbf{p}^{k_q}(\mathbf{w}_i)||_2^{2} = ||\mathbf{O}-\mathbf{Q}||_2^2\\
\end{array}
\end{equation}
This generalization has the advantage of being faster to compute since it treats the concatenated permutation vector as a whole. Moreover, it does not
require square roots and it can be evaluated using the cosine. 
Defining in the same way as above:
\begin{equation}
\begin{array}{ll}
\mathbf{\widehat{p}}^{k_x}(\mathbf{v}_i)=\mathbf{k_x}-\mathbf{p}^{k_x}(\mathbf{v}_i)\\
\mathbf{\widehat{p}}^{k_q}(\mathbf{w}_i)=\mathbf{k_q}-\mathbf{p}^{k_q}(\mathbf{w}_i).
\end{array} \label{eq:minusvectsvlad}
\end{equation}
By a similar procedure shown above, it is possible to prove that also in this case $sim_{cos} (\mathbf{V}, \mathbf{W})\propto \alpha - \beta\, \tilde{d}^2(\mathbf{V}, \mathbf{W}) $ holds.

In order to correctly match the transformed blockwise vectors, we need to extended the reference dictionary to distinguish the key produced from sub-vectors $\mathbf{v}_i$ with different subscript $i$. There for a set $m$ of reference objects, and $\mathcal{K}$ element in the VLAD codebook, we employ dictionary including a set of $m \times \mathcal{K}$ keys $\tau_{i,j}$ ($1\leq i \leq m$, $1\leq j \leq \mathcal{K}$).

For example, we associate, say, the set of keys A$_1$, B$_1$,... to the sub-vector $\mathbf{v}_1$,  A$_2$, B$_2$,... to the sub-vector $\mathbf{v}_2$, and so on.

\subsection{Dealing with VLAD Ambiguities}
One of the well-known problems of VLAD happens when no local descriptor is assigned to a codeword \cite{peng2014}. A simple approach to this problem is produce a sub-vector of all zeros ($\mathbf{v}_i=\mathbf{0}$) but this has the disadvantage to be ambiguous since it is identical to the case in which the mean of the local descriptors assigned to a codeword is equal to the codeword itself.

Moreover, as pointed out by \cite{spyromitros2014comprehensive}, given two images and the corresponding VLAD vectors $\mathbf{V}$ and $\mathbf{W}$, and assuming that $\mathbf{v}_i=\mathbf{0}$, the contribution of codeword $\bm{\mu}_i$ to the cosine similarity of $\mathbf{V}$ and $\mathbf{W}$ will be the same when either $\mathbf{w}_i=\mathbf{0}$ or $\mathbf{w}_i\neq\mathbf{0}$. Therefore, this under-estimates the importance of jointly zero components, which gives some limited yet important evidence on visual similarity \cite{jegou2012negative}. In \cite{jegou2012negative}, this problem was treated by measuring the cosine between vectors $\mathbf{V}$ and $\mathbf{W}$ at different point from the origin.

This technique, however, did not lead to significant improvement of our experiments. To tackle this problem, we simply get rid of the sub-vectors $\mathbf{v}_i=\mathbf{0}$ and omit to transform them in text. Mathematically, this means that we assume $\mathbf{\widehat{p}}^{k_x}(\mathbf{0}) = \mathbf{0}$.



\section{Experiments}
\label{sec:experiments} \noindent

\subsection{Setup}
\label{setup}

INRIA Holidays \cite{JegouCVPR2010,jegou12} is a collection of 1,491 holiday images. The authors selected 500 queries and for each of them a list of positive results.
As in \cite{Jegou2009,Jegou2010,jegou12}, to evaluate the approaches on a large scale, we merged the Holidays dataset with the Flickr1M collection\footnote{\url{http://press.liacs.nl/mirflickr/}}.
SIFT features have been extracted by Jegou et al. for both the Holidays and the Flickr1M datasets\footnote{\url{http://lear.inrialpes.fr/~jegou/data.php}}.

For representing the images using the VLAD approach, we selected 64 reference features using \emph{k-means} over a subset of the Flickr1M dataset.

All experiments were conducted on a Intel Core i7 CPU, 2.67 GHz with 12.0 GB of RAM a 2TB 7200 RPM HD for the Lucene index. We used Lucene v4.7 running on Java 6 64 bit.

The quality of the retrieved images is typically evaluated by means of precision and recall measures. As in many other papers \cite{Jegou2009,Perronnin2010,jegou12}, we combined this information by means of the mean Average Precision (mAP), which represents the area below the precision and recall curve.

\subsection{Results}

In a first experimental analysis, we compared the performance of blockwise approach versus the baseline approach (with and without reordering) that threats the VLAD vectors as whole-objects, which was studied in \cite{Amato:2014:IVL:2578726.2578788}. In this latter approach, as explained Section \ref{sec:proposed approach}, since the performance was low, we had to reorder the best results using the actual distance between the VLAD descriptors. With this experiment, we want to show that with the blockwise approach this phase is no longer necessary, and the search is only based on the result provided by text-search engine Lucene. For the baseline approach, we used $m=$4,000 reference objects while for blockwise, 20,000. In both cases, we set $k_x = 50$, which, we recall, is the number of closest reference objects used during indexing.

\begin{figure}[t]
    \centering
    	\includegraphics[ trim=27mm 15mm 27mm 20mm, width=0.98\columnwidth]{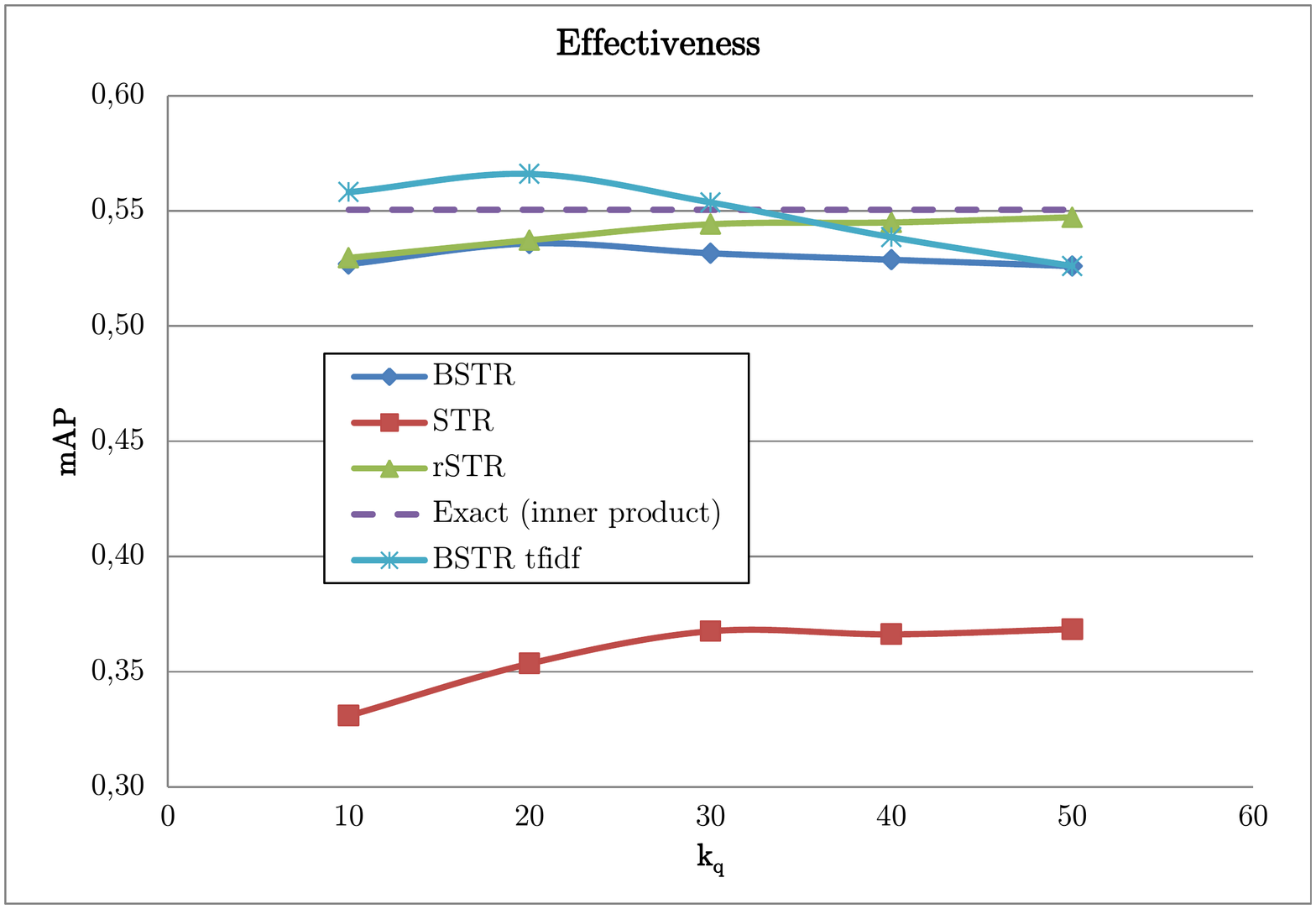}
    \caption{Effectiveness (mAP) of the various approach for the INRIA Holidays dataset, using $k_x=50$ for STR, rSTR, BSTR, and BSTR {tfidf} (higher
    	values mean better reults).}
    \label{fig:basevsblock}
\end{figure}

Figure \ref{fig:basevsblock} shows this comparison in terms of mAP. We refer to baseline approach as STR, the baseline approach with reordering as rSTR, and to blockwise approach as BSTR. For the rSTR approach, we reordered the first 1,000 objects of the results set. The horizontal line at the top represents the performance obtained matching the original VLAD descriptors with the inner product, performing a sequential scan of the dataset, which exhibits a mAP of 0.55. The graph in the middle shows the mAP of our approach (BSTR) versus the baseline approach without reordering (STR) and with reordering (rSTR). The graphs show also how the performance changes varying $k_q$ (the number of closest reference objects for the query) from 10 to 50.

An interesting by-product of the experiment is that, we obtain a little improvement of the mAP for the BSTR approach when the number of reference objects used for the query is 20.

A quite intuitive way of generalizing the idea of reducing the size of the query is to exploit the knowledge of the \emph{tf*idf} (i.e., term frequency * inverse document frequency) statistic of the BSTR textual representation. Instead of simply reducing the $k_q$ of the query, i.e., the top-$k_q$ element nearest to the query, we can retain the elements that exhibit greater values of \emph{tf*idf} starting from the document generated with $k_q=50$ and eliminate the others. Therefore, we take, for instance, the first 40 elements that have best \emph{tf*idf}, the first 30 elements, and so on. Figure \ref{fig:basevsblock} shows the performance of this approach, with the name `BSTR tfidf'. It is interesting to note that we had not only an important  improvement of the mAP for increasing reduction of the queries but also that this approach outperforms the performance of the inner product on the original VLAD dataset.

In order to ascertain the soundness of the proposed approach, we tested it on the larger and challenger Flickr1M dataset.

\begin{figure}[t]
    \centering
    	\includegraphics[ trim=27mm 15mm 27mm 20mm, width=0.98\columnwidth]{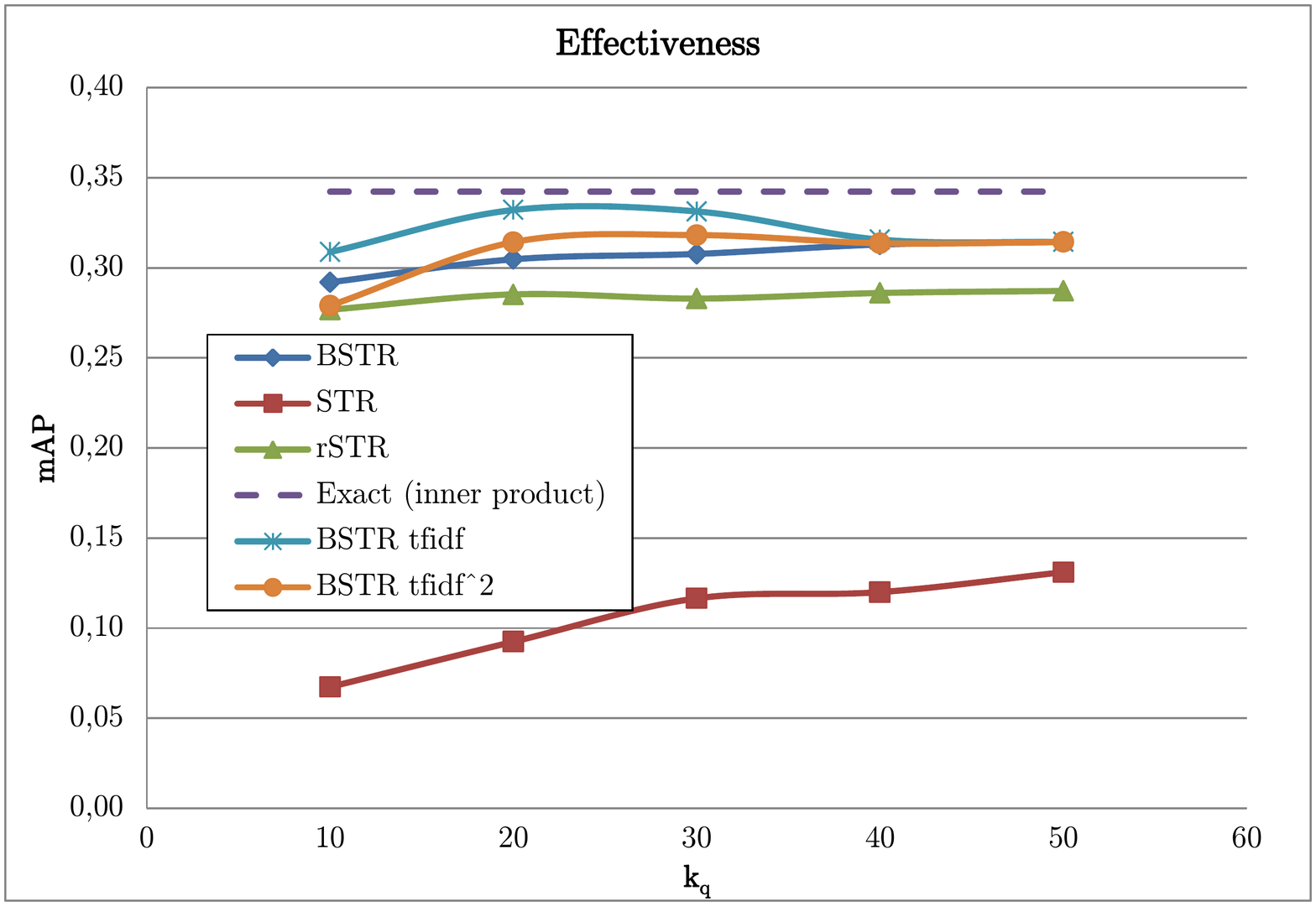}
    \caption{Effectiveness (mAP) of the various approach for the INRIA Holidays + Flickr1M dataset, using $k_x=50$ for STR, rSTR, BSTR, and BSTR tfidf. While for BSTR tfidf$^2$, we set $k_x=k_q$  (higher
    	values mean better reults).}
    \label{fig:basevsblock1M}
\end{figure}

The results are shown in Figure \ref{fig:basevsblock1M}. We can see that BSTR tfidf is still the winner in terms of mAP. However, in this case all the techniques exhibit lower performance with respect the inner product on the original VLAD dataset. The latter test is performed as a sequential scan of the entire dataset obtaining a mAP of 0.34. The results presented in this figure also show the performance of the approach called BSTR tfidf$^2$, which consists in applying the reduction of the blockwise textual representation using \emph{tf*idf} also for the indexed document (in addition to the queries), setting $k_x = k_q$ for all the experiments. The mAPs values in this case are slightly lower than BSTR tfidf, however, as we are going to see in the next experiment there is a great advance in terms of space occupation.

\begin{figure}[t]
    \centering
    	\includegraphics[ trim=27mm 15mm 27mm 20mm, width=0.98\columnwidth]{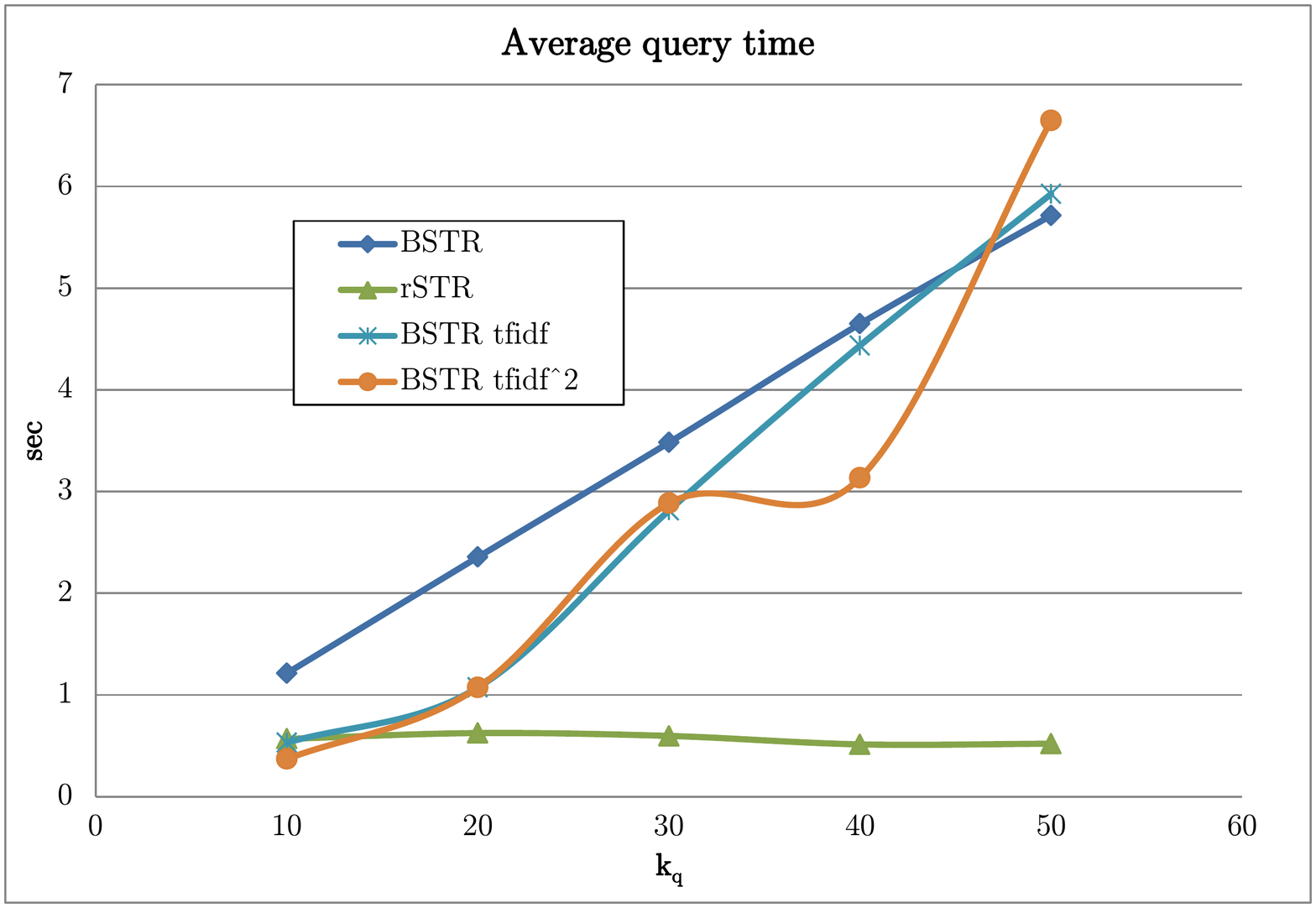}
    \caption{Average time per query in seconds of the various approaches for the INRIA Holidays + Flickr1M dataset, using $k_x=50$ for rSTR, BSTR, and BSTR tfidf. While for BSTR tfidf$^2$, we set $k_x=k_q$  (higher values mean worse performance).}
    \label{fig:basevsblock1Mtimes}
\end{figure}

In order to assess which approach is most promising, we have also evaluated the efficiency in terms of space and time overhead. Figure \ref{fig:basevsblock1Mtimes} shows the average time for a query for the proposed approaches. The rSTR approach considers also the time for reordering the result set, however, its average time is obtained using a solid state disk (SSD) disk in which the original VLAD vectors are available for the reordering. The SSD is necessary to guarantee fast random I/O, while using a standard disk the seek time would affect the query time of more than one order of magnitude.

\begin{figure}[t]
    \centering
    	\includegraphics[ trim=26mm 15mm 26mm 20mm, width=0.98\columnwidth]{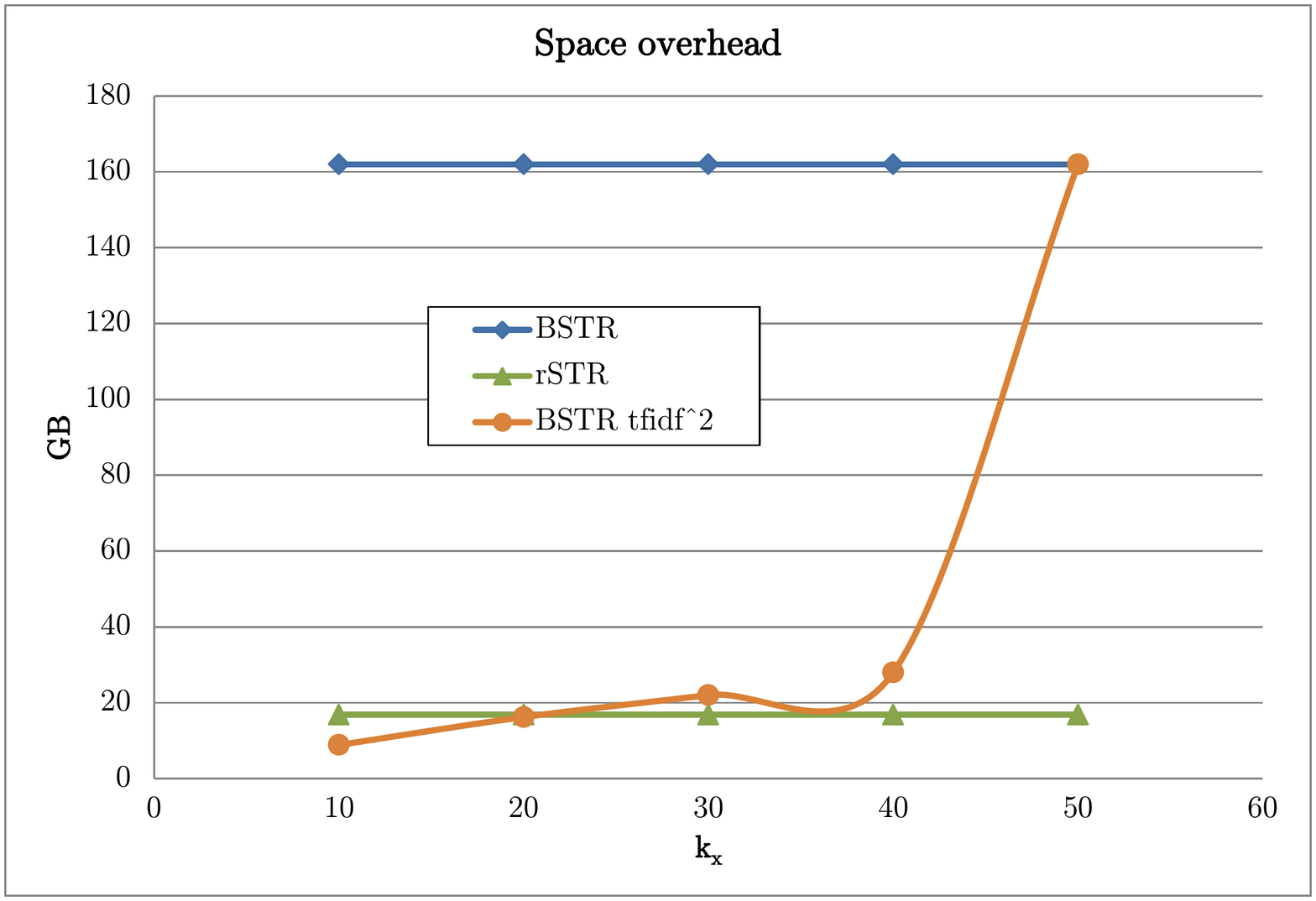}
    \caption{Space occupation of the index for the different type of solutions, using the same value of $k_x=50$ for BSTR and rSTR, and varing $k_x$ for BSTR tfidf$^2$. Note that for the rSTR, we consider also the overhead for the storage of the VLAD vectors used for the reordering of the results (higher values mean greater occupations).}
    \label{fig:basevsblock1Mspace}
\end{figure}

Figure \ref{fig:basevsblock1Mspace} presents the index occupation expressed in GB. The rSTR approach occupies 16.8 GB on the disk, including the overhead for the storage of the VLAD vectors used for the reordering of the results. The BSTR tfidf$^2$ solution has great impact of the space occupation: just for a reduction of the 20\% of the documents ( i.e., from $k_x=50$ to $k_x=40$) we get a reduction of the 80\% for the inverted file.

Considering all the alternatives seen so far, an optimal choice could be BSTR tfidf$^2$ with $k_x=k_q=20$, which is efficient in term of both time and space overheads and still maintains satisfactory mAP.

\section{Conclusions and Future Work}
\label{sec:conclusions} \noindent
In this work, we proposed a `blockwise' extension of surrogate text representation, which is in principle applicable not only to VLAD but also to any other vector or compound metric objects. The main advantage of this approach is the elimination for the need of the reordering phase.
Using the same hardware and text search engine (i.e., Lucene), we were able to compare with the state-of-the-art baseline STR approach exploiting the reordering phase.

The experimental evaluation on the blockwise extension revealed very promising performance in terms of mAP and response time. However, the drawback of it resides in the expansion of the number of terms in the  textual representation of the VLADs. This produces an inverted index that, using Lucene, is one order of magnitude greater than the baseline STR. To alleviate this problem, we propose to shrink the index reducing the document, as we did for the query, by eliminating the terms associated with a low value of \emph{tf*idf} weight.
This approach is very effective but has the disadvantage that need a double indexing phase or at least a pre-analysis of the dataset in order to calculate the \emph{tf*idf} weight of the terms. Future work will investigate this aspect in more detail.

\section*{\uppercase{Acknowledgements}}
This work was partially supported by EAGLE, Europeana network of Ancient Greek and Latin Epigraphy, co-founded by the European Commision, CIP-ICT-PSP.2012.2.1 - Europeana and creativity, Grant Agreement n. 325122.

\bibliographystyle{apalike}
{\small
    \bibliography{bib}}

\begin{thebibliography}{}

\bibitem[Amato et~al., 2013a]{amato13}
Amato, G., Bolettieri, P., Falchi, F., and Gennaro, C. (2013a).
\newblock Large scale image retrieval using vector of locally aggregated
  descriptors.
\newblock In Brisaboa, N., Pedreira, O., and Zezula, P., editors, {\em
  Similarity Search and Applications}, volume 8199 of {\em Lecture Notes in
  Computer Science}, pages 245--256. Springer Berlin Heidelberg.

\bibitem[Amato et~al., 2011]{AmatoCBMI2011}
Amato, G., Bolettieri, P., Falchi, F., Gennaro, C., and Rabitti, F. (2011).
\newblock Combining local and global visual feature similarity using a text
  search engine.
\newblock In {\em Content-Based Multimedia Indexing (CBMI), 2011 9th
  International Workshop on}, pages 49 --54.

\bibitem[Amato et~al., 2013b]{amato13:onReducing}
Amato, G., Falchi, F., and Gennaro, C. (2013b).
\newblock On reducing the number of visual words in the bag-of-features
  representation.
\newblock In {\em {VISAPP} 2013 - Proceedings of the International Conference
  on Computer Vision Theory and Applications}, volume~1, pages 657--662.

\bibitem[Amato et~al., 2014a]{Amato:2014:IVL:2578726.2578788}
Amato, G., Falchi, F., Gennaro, C., and Bolettieri, P. (2014a).
\newblock Indexing vectors of locally aggregated descriptors using inverted
  files.
\newblock In {\em Proceedings of International Conference on Multimedia
  Retrieval}, ICMR '14, pages 439:439--439:442.

\bibitem[Amato et~al., 2014b]{mtap12}
Amato, G., Gennaro, C., and Savino, P. (2014b).
\newblock {MI-File}: using inverted files for scalable approximate similarity
  search.
\newblock {\em Multimedia Tools and Applications}, 71(3):1333--1362.

\bibitem[Arandjelovic and Zisserman, 2013]{arandjelovic13:allAbVALD}
Arandjelovic, R. and Zisserman, A. (2013).
\newblock All about {VLAD}.
\newblock In {\em Computer Vision and Pattern Recognition (CVPR), 2013 IEEE
  Conference on}, pages 1578--1585.

\bibitem[Bay et~al., 2006]{bay06}
Bay, H., Tuytelaars, T., and Van~Gool, L. (2006).
\newblock {SURF: Speeded Up Robust Features}.
\newblock In Leonardis, A., Bischof, H., and Pinz, A., editors, {\em Computer
  Vision - ECCV 2006}, volume 3951 of {\em Lecture Notes in Computer Science},
  pages 404--417. Springer Berlin Heidelberg.

\bibitem[Boureau et~al., 2010]{boureau10}
Boureau, Y.-L., Bach, F., LeCun, Y., and Ponce, J. (2010).
\newblock Learning mid-level features for recognition.
\newblock In {\em Computer Vision and Pattern Recognition (CVPR), 2010 IEEE
  Conference on}, pages 2559--2566.

\bibitem[Chavez et~al., 2008]{Navarro2008}
Chavez, G., Figueroa, K., and Navarro, G. (2008).
\newblock Effective proximity retrieval by ordering permutations.
\newblock {\em Pattern Analysis and Machine Intelligence, IEEE Transactions
  on}, 30(9):1647 --1658.

\bibitem[Chen et~al., 2011]{chen11}
Chen, D., Tsai, S., Chandrasekhar, V., Takacs, G., Chen, H., Vedantham, R.,
  Grzeszczuk, R., and Girod, B. (2011).
\newblock Residual enhanced visual vectors for on-device image matching.
\newblock In {\em Signals, Systems and Computers (ASILOMAR), 2011 Conference
  Record of the Forty Fifth Asilomar Conference on}, pages 850--854.

\bibitem[Chum et~al., 2007]{chum07}
Chum, O., Philbin, J., Sivic, J., Isard, M., and Zisserman, A. (2007).
\newblock Total recall: Automatic query expansion with a generative feature
  model for object retrieval.
\newblock In {\em Computer Vision, 2007. ICCV 2007. IEEE 11th International
  Conference on}, pages 1--8.

\bibitem[Csurka et~al., 2004]{csurka04}
Csurka, G., Dance, C., Fan, L., Willamowski, J., and Bray, C. (2004).
\newblock Visual categorization with bags of keypoints.
\newblock {\em Workshop on statistical learning in computer vision, ECCV},
  1(1-22):1--2.

\bibitem[Datar et~al., 2004]{Datar:2004}
Datar, M., Immorlica, N., Indyk, P., and Mirrokni, V.~S. (2004).
\newblock Locality-sensitive hashing scheme based on p-stable distributions.
\newblock In {\em Proceedings of the twentieth annual symposium on
  Computational geometry}, SCG '04, pages 253--262.

\bibitem[Delhumeau et~al., 2013]{delhumeau13}
Delhumeau, J., Gosselin, P.-H., J{\'e}gou, H., and P{\'e}rez, P. (2013).
\newblock Revisiting the {VLAD} image representation.
\newblock In {\em Proceedings of the 21st ACM International Conference on
  Multimedia}, MM '13, pages 653--656.

\bibitem[Esuli, 2009]{MiPai}
Esuli, A. (2009).
\newblock {MiPai: Using the PP-Index to Build an Efficient and Scalable
  Similarity Search System}.
\newblock In {\em Proceedings of the 2009 Second International Workshop on
  Similarity Search and Applications}, SISAP '09, pages 146--148.

\bibitem[Fagin et~al., 2003]{FAGIN2003b}
Fagin, R., Kumar, R., and Sivakumar, D. (2003).
\newblock Comparing top-k lists.
\newblock {\em SIAM J. of Discrete Math.}, 17(1):134--160.

\bibitem[Gennaro et~al., 2010]{ecdl10}
Gennaro, C., Amato, G., Bolettieri, P., and Savino, P. (2010).
\newblock An approach to content-based image retrieval based on the lucene
  search engine library.
\newblock In Lalmas, M., Jose, J., Rauber, A., Sebastiani, F., and Frommholz,
  I., editors, {\em Research and Advanced Technology for Digital Libraries},
  volume 6273 of {\em Lecture Notes in Computer Science}, pages 55--66.
  Springer Berlin Heidelberg.

\bibitem[Jaakkola and Haussler, 1998]{jaakkola98}
Jaakkola, T. and Haussler, D. (1998).
\newblock Exploiting generative models in discriminative classifiers.
\newblock In {\em In Advances in Neural Information Processing Systems 11},
  pages 487--493.

\bibitem[J{\'e}gou and Chum, 2012]{jegou2012negative}
J{\'e}gou, H. and Chum, O. (2012).
\newblock Negative evidences and co-occurences in image retrieval: The benefit
  of pca and whitening.
\newblock In Fitzgibbon, A., Lazebnik, S., Perona, P., Sato, Y., and Schmid,
  C., editors, {\em Computer Vision--ECCV 2012}, volume 7573 of {\em Lecture
  Notes in Computer Science}, pages 774--787. Springer.

\bibitem[J{\'e}gou et~al., 2008]{jegou08}
J{\'e}gou, H., Douze, M., and Schmid, C. (2008).
\newblock Hamming embedding and weak geometric consistency for large scale
  image search.
\newblock In Forsyth, D., Torr, P., and Zisserman, A., editors, {\em Computer
  Vision -- ECCV 2008}, volume 5302 of {\em Lecture Notes in Computer Science},
  pages 304--317. Springer Berlin Heidelberg.

\bibitem[J{\'e}gou et~al., 2009]{Jegou2009}
J{\'e}gou, H., Douze, M., and Schmid, C. (2009).
\newblock Packing bag-of-features.
\newblock In {\em Computer Vision, 2009 IEEE 12th International Conference on},
  pages 2357 --2364.

\bibitem[J\'{e}gou et~al., 2010]{Jegou2010}
J\'{e}gou, H., Douze, M., and Schmid, C. (2010).
\newblock Improving bag-of-features for large scale image search.
\newblock {\em International Journal of Computer Vision}, 87:316--336.

\bibitem[J{\'e}gou et~al., 2010a]{JegouCVPR2010}
J{\'e}gou, H., Douze, M., Schmid, C., and P\'erez, P. (2010a).
\newblock Aggregating local descriptors into a compact image representation.
\newblock In {\em IEEE Conference on Computer Vision \& Pattern Recognition},
  pages 3304--3311.

\bibitem[J{\'e}gou et~al., 2012]{jegou12}
J{\'e}gou, H., Perronnin, F., Douze, M., S{\`a}nchez, J., P{\'e}rez, P., and
  Schmid, C. (2012).
\newblock Aggregating local image descriptors into compact codes.
\newblock {\em IEEE Transactions on Pattern Analysis and Machine Intelligence},
  34(9):1704--1716.

\bibitem[J{\'e}gou et~al., 2010b]{jegou10:AccurateImage}
J{\'e}gou, H., Schmid, C., Harzallah, H., and Verbeek, J. (2010b).
\newblock Accurate image search using the contextual dissimilarity measure.
\newblock {\em Pattern Analysis and Machine Intelligence, IEEE Transactions
  on}, 32(1):2--11.

\bibitem[Lazebnik et~al., 2006]{lazebnik06}
Lazebnik, S., Schmid, C., and Ponce, J. (2006).
\newblock Beyond bags of features: Spatial pyramid matching for recognizing
  natural scene categories.
\newblock In {\em Computer Vision and Pattern Recognition, 2006 IEEE Computer
  Society Conference on}, volume~2.

\bibitem[Lowe, 2004]{lowe04}
Lowe, D. (2004).
\newblock Distinctive image features from scale-invariant keypoints.
\newblock {\em International Journal of Computer Vision}, 60(2):91--110.

\bibitem[McLachlan and Peel, 2000]{mclachlan2000}
McLachlan, G. and Peel, D. (2000).
\newblock {\em Finite Mixture Models}.
\newblock Wiley series in probability and statistics. Wiley.

\bibitem[Peng et~al., 2014]{peng2014}
Peng, X., Wang, L., Qiao, Y., and Peng, Q. (2014).
\newblock Boosting vlad with supervised dictionary learning and high-order
  statistics.
\newblock In Fleet, D., Pajdla, T., Schiele, B., and Tuytelaars, T., editors,
  {\em Computer Vision - ECCV 2014}, volume 8691 of {\em Lecture Notes in
  Computer Science}, pages 660--674. Springer International Publishing.

\bibitem[Perd'och et~al., 2009]{perdoch09}
Perd'och, M., Chum, O., and Matas, J. (2009).
\newblock Efficient representation of local geometry for large scale object
  retrieval.
\newblock In {\em Computer Vision and Pattern Recognition, 2009. CVPR 2009.
  IEEE Conference on}, pages 9--16.

\bibitem[Perronnin and Dance, 2007]{perronnin07}
Perronnin, F. and Dance, C. (2007).
\newblock Fisher kernels on visual vocabularies for image categorization.
\newblock In {\em Computer Vision and Pattern Recognition, 2007. CVPR '07. IEEE
  Conference on}, pages 1--8.

\bibitem[Perronnin et~al., 2010]{Perronnin2010}
Perronnin, F., Liu, Y., Sanchez, J., and Poirier, H. (2010).
\newblock Large-scale image retrieval with compressed fisher vectors.
\newblock In {\em Computer Vision and Pattern Recognition (CVPR), 2010 IEEE
  Conference on}, pages 3384 --3391.

\bibitem[Philbin et~al., 2007]{philbin07}
Philbin, J., Chum, O., Isard, M., Sivic, J., and Zisserman, A. (2007).
\newblock Object retrieval with large vocabularies and fast spatial matching.
\newblock In {\em Computer Vision and Pattern Recognition, 2007. CVPR 2007.
  IEEE Conference on}, pages 1--8.

\bibitem[Philbin et~al., 2008]{philbin08}
Philbin, J., Chum, O., Isard, M., Sivic, J., and Zisserman, A. (2008).
\newblock Lost in quantization: Improving particular object retrieval in large
  scale image databases.
\newblock In {\em Computer Vision and Pattern Recognition, 2008. CVPR 2008.
  IEEE Conference on}, pages 1--8.

\bibitem[Salton and McGill, 1986]{salton86}
Salton, G. and McGill, M.~J. (1986).
\newblock {\em Introduction to Modern Information Retrieval}.
\newblock McGraw-Hill, Inc., New York, NY, USA.

\bibitem[Sivic and Zisserman, 2003]{sivic03}
Sivic, J. and Zisserman, A. (2003).
\newblock Video google: A text retrieval approach to object matching in videos.
\newblock In {\em Proceedings of the Ninth IEEE International Conference on
  Computer Vision - Volume 2}, ICCV '03, pages 1470--1477.

\bibitem[Spyromitros-Xioufis et~al., 2014]{spyromitros2014comprehensive}
Spyromitros-Xioufis, E., Papadopoulos, S., Kompatsiaris, I.~Y., Tsoumakas, G.,
  and Vlahavas, I. (2014).
\newblock A comprehensive study over vlad and product quantization in
  large-scale image retrieval.
\newblock {\em Multimedia, IEEE Transactions on}, 16(6):1713--1728.

\bibitem[Thomee et~al., 2010]{thomee10}
Thomee, B., Bakker, E.~M., and Lew, M.~S. (2010).
\newblock {TOP-SURF}: A visual words toolkit.
\newblock In {\em Proceedings of the International Conference on Multimedia},
  MM '10, pages 1473--1476.

\bibitem[Tolias and Avrithis, 2011]{tolias11:SpeededUp}
Tolias, G. and Avrithis, Y. (2011).
\newblock Speeded-up, relaxed spatial matching.
\newblock In {\em Computer Vision (ICCV), 2011 IEEE International Conference
  on}, pages 1653--1660.

\bibitem[Tolias and J{\'e}gou, 2013]{tolias13:queryExp}
Tolias, G. and J{\'e}gou, H. (2013).
\newblock {Local visual query expansion: Exploiting an image collection to
  refine local descriptors}.
\newblock Research Report RR-8325, {INRIA}.

\bibitem[Van~Gemert et~al., 2010]{vanGemert10}
Van~Gemert, J., Veenman, C., Smeulders, A., and Geusebroek, J.-M. (2010).
\newblock Visual word ambiguity.
\newblock {\em Pattern Analysis and Machine Intelligence, IEEE Transactions
  on}, 32(7):1271--1283.

\bibitem[Van~Gemert et~al., 2008]{vanGemert08}
Van~Gemert, J.~C., Geusebroek, J.-M., Veenman, C.~J., and Smeulders, A.~W.
  (2008).
\newblock Kernel codebooks for scene categorization.
\newblock In Forsyth, D., Torr, P., and Zisserman, A., editors, {\em Computer
  Vision - ECCV 2008}, volume 5304 of {\em Lecture Notes in Computer Science},
  pages 696--709. Springer Berlin Heidelberg.

\bibitem[Wang et~al., 2010]{wang10}
Wang, J., Yang, J., Yu, K., Lv, F., Huang, T., and Gong, Y. (2010).
\newblock Locality-constrained linear coding for image classification.
\newblock In {\em Computer Vision and Pattern Recognition (CVPR), 2010 IEEE
  Conference on}, pages 3360--3367.

\bibitem[Witten et~al., 1999]{witten99}
Witten, I.~H., Moffat, A., and Bell, T.~C. (1999).
\newblock {\em Managing gigabytes: compressing and indexing documents and
  images}.
\newblock Multimedia Information and Systems Series. Morgan Kaufmann
  Publishers.

\bibitem[Yang et~al., 2009]{yang09}
Yang, J., Yu, K., Gong, Y., and Huang, T. (2009).
\newblock Linear spatial pyramid matching using sparse coding for image
  classification.
\newblock In {\em Computer Vision and Pattern Recognition, 2009. CVPR 2009.
  IEEE Conference on}, pages 1794--1801.

\bibitem[Zezula et~al., 2006]{metric-book}
Zezula, P., Amato, G., Dohnal, V., and Batko, M. (2006).
\newblock {\em Similarity Search: The Metric Space Approach}, volume~32 of {\em
  Advances in Database Systems}.
\newblock Springer-Verlag.

\bibitem[Zhao et~al., 2013]{zhao13}
Zhao, W.-L., J{\'e}gou, H., and Gravier, G. (2013).
\newblock {Oriented pooling for dense and non-dense rotation-invariant
  features}.
\newblock In {\em {BMVC - 24th British Machine Vision Conference}}.

\end{thebibliography}

\end{document}